\documentclass[conference]{IEEEtran}
\usepackage{cite}
\usepackage{amsmath,amssymb,amsfonts}
\usepackage{multirow}
\usepackage{algorithmic}
\usepackage{graphicx}
\usepackage{textcomp}
\usepackage{subcaption}
\usepackage{pgfplots}
\pgfplotsset{compat=1.14}
\usepackage{tikz}
\usepgfplotslibrary{units}
\usepackage{hyperref}
\usepackage{url}
\usepackage{xcolor}
\def\BibTeX{{\rm B\kern-.05em{\sc i\kern-.025em b}\kern-.08em
    T\kern-.1667em\lower.7ex\hbox{E}\kern-.125emX}}

\begin{document}
\title{Optimizing Deep Neural Networks with Multiple Search Neuroevolution}

\author{\IEEEauthorblockN{Ahmed Aly}
\IEEEauthorblockA{\textit{Computer Engineering} \\
\textit{University of Virginia}\\
Charlottesville, VA \\
aaa2cn@virginia.edu}
\and
\IEEEauthorblockN{David Weikersdorfer}
\IEEEauthorblockA{\textit{NVIDIA Robotics} \\
Santa Clara, CA \\
dweikersdorfer@nvidia.com}
\and
\IEEEauthorblockN{Claire Delaunay}
\IEEEauthorblockA{\textit{NVIDIA Robotics} \\
Santa Clara, CA \\
cdelaunay@nvidia.com}
}

\maketitle

\begin{abstract}
This paper presents an evolutionary metaheuristic called Multiple Search Neuroevolution (MSN) to optimize deep neural networks. The algorithm attempts to search multiple promising regions in the search space simultaneously, maintaining sufficient distance between them. It is tested by training neural networks for two tasks, and compared with other optimization algorithms. The first task is to solve Global Optimization functions with challenging topographies. We found to MSN to outperform classic optimization algorithms such as Evolution Strategies, reducing the number of optimization steps performed by at least 2X.

The second task is to train a convolutional neural network (CNN) on the popular MNIST dataset. Using 3.33\% of the training set, MSN reaches a validation accuracy of 90\%. Stochastic Gradient Descent (SGD) was able to match the same accuracy figure, while taking 7X less optimization steps. Despite lagging, the fact that the MSN metaheurisitc trains a 4.7M-parameter CNN suggests promise for future development. This is by far the largest network ever evolved using a pool of only 50 samples.
\end{abstract}


\section{Introduction}
In recent years, deep neural networks have been employed for tasks in various domains like Object Detection \cite{Liu2016Ssd:Detector}, Robotic Grasping \cite{Lenz2015DeepGrasps} and Machine Translation \cite{Bahdanau2014NeuralTranslate}. They are quite popular and powerful given their representational capacity and automatic feature extraction. To train those networks the standard algorithms used are Gradient Descent \cite{CauchyMethodeSimultanees} variations such as Stochastic Gradient Descent (SGD) \cite{Robbins1951AMethod} and ADAM \cite{Kingma2014Adam:Optimization}, all employing Backpropagation \cite{WERBOS1974BeyondScience}. Remarkably, Gradient Descent, SGD and Backpropagation were reported in 1847, 1951 and 1974 respectively. Since then, many techniques and variations have been reported over the years such as ADAM and Batch Normalization \cite{Ioffe2015BatchShift} in 2015.

In general, gradient-based algorithms are preferred when training neural networks because of their ability to handle complexities introduced by networks with millions of weights. However, there are also limitations when using a gradient-based optimization algorithm such as SGD. The neural network architecture and the loss function have to be end-to-end differentiable. Consequently, there are a number of problems that can not be directly modeled or addressed without some alterations such as Formal Logic, Discrete Action and Hard Attention \cite{Luong2015EffectiveTranslation} \cite{Xu2015ShowAttention}. Another limitation is the lack of exploration due to the greedy nature of the algorithm, i.e. gradient-following. This makes the algorithm somewhat linear in the discovered solutions due to little exploration of the search space. For tasks that require exploration, this can make the training process challenging. Furthermore, the algorithm is prone to getting stuck in local minima and saddle points. While there are remedies to such challenges, e.g. using Dropout, they may not always produce the desired effect.

For those reasons, we ventured to investigate derivative-free optimization algorithms that can potentially train deep neural networks. Optimization algorithms generally are divided into two categories, exact and heuristic \cite{Sorensen2015Metaheuristics-theExposed}. Due to the combinatorial complexities of training neural networks, heuristics are almost always used. They are neither guaranteed to converge nor reach an optimal solution. Heuristics can further be divided into two categories, single-solution and population-based. Otherwise, there are many more lines to distinguish different families of algorithms by. Single-solution heuristics, also called trajectory methods, attempt to iteratively improve upon a single candidate solution. Single-solution heuristics are generally exploitation-based algorithms \cite{2005ParallelMetaheuristics}. On the other hand, population-based heuristics attempt to improve a "population" (set) of solutions based on their "fitness" (performance) according to an objective function, over generations (optimization steps). Population-based heuristics are generally exploration-based. For those algorithms to achieve a solution, there must be some differentiator between the populations throughout the optimization process.

Particular to neural networks are Neuroevolution algorithms, a derivative of evolutionary computation algorithms. In 1994 Ronald and Schoenauer first reported using genetic algorithms to optimize a simple neural network for a toy task \cite{Ronald1994GeneticControl}. Since then there were many implementations and variations on Neuroevolution, such as NEAT in 2002 \cite{Miikkulainen2002EvolvingTopologies}. In general, Neuroevolution algorithms can be split into two categories. The first, such as in NEAT, attempts to evolve the topology along with the weights of the network. The second attempts to evolve only the weights, such as in \cite{Such2017DeepLearning}. Since evolving the topology would add another aspect of complexity, we are only concerned with evolving the weights of the network. In addition, evolving the network topology does not allow a somewhat direct comparison with SGD.

To that end, this paper reports a metaheuristic called Multiple Search Neuroevolution (MSN). A metaheuristic is a "strategy" that guides the search process, not an algorithm for a particular problem \cite{2005ParallelMetaheuristics}. We report the MSN metaheuristic as a set of principles that shape an overall guide strategy. It tries to search multiple regions of the search space, while maintaining a certain distance between those regions to ensure diversity. Despite searching multiple regions simultaneously, it is a serial algorithm where populations (or samples as we will later call them) are generated and evaluated one at a time. Throughout this paper the abbreviation MSN and the terms Our Algorithm \& Metaheuristic will be used interchangeably, depending on context, to refer to the proposed metaheuristic.

\section{Background}
\subsection{A problem of Spaces}
Training neural networks is essentially an iterative optimization process. The optimization parameters and objective function are the network's weights and loss function respectively. The optimization process aims to find a desirable set of weights given target input/output pairs such as in \cite{Goodfellow2014GenerativeNets}, \cite{Finn2016DeepLearning}, \cite{Mnih2013PlayingLearning}. Through the weights, a mapping between the given input and the desired output is created. The weights are adjusted in order to approximate a function that achieves said mapping. The larger or deeper the network, generally, the greater its capacity to approximate more complex functions. This is referred to as Representational Capacity in \cite{Goodfellow2016DeepLearning}. Increasing representational capacity allows the creation of more complex mappings, i.e. approximation of higher-order functions. In turn, the network can be used to address more challenging tasks.

In this context, it is important to examine the abstract concepts of Spaces. The \textbf{search space} is the space (set) of all possible weight configurations. The wider and deeper the network, the larger the search space. The type of neuron itself also affects search space. In a recurrent neural network, for instance, there is an additional recurrent parameter to learn. Besides architecture, another important factor is the range of the weights. That is, binary weights would lead to a considerably smaller search space than 32-bit representations (FP32). Weights in a neural network are independent. Each weight is a free-valued parameter, constrained only by the range of representation.

The \textbf{observation space} is the space (set) of all possible observations. Consider the case of the MNIST image dataset. The input size is a 28x28x1 full-precision (FP32) image (matrix). The entire dataset consists of 70,000 such images. The observation space, however, is made of all possible 28x28x1 full-precision images (matrices). Thus the 70,000 images are not the entire observation space. They are simply a small fraction of it.

The \textbf{solution space} can be defined as the space (set) of all possible outcomes of the network. If the network only has one binary output neuron, for example, then the entire solution space consists only of two members [0, 1]. Another example would be Generative Adversarial Networks that generate images. The solution space is the set of all possible images (matrices). The larger the image to be generated, the larger the solution space, and generally the more difficult it is to find a high-quality solution.

Finally, there is \textbf{loss space} or gradient space. This is the space (set) of all possible error/cost values of the objective (loss) function. This is the space that the solver has to traverse, in order to look for better solution candidate(s) in the the search space. The topography of this space can sometimes be quite challenging to navigate. It may not be smooth, continuous, convex or even information-bearing. Salimans, Ho, Chen, Sido and Sutskever report this in \cite{Salimans2017EvolutionLearning} when they refer to cases where the gradients are "uninformative". Furthermore, the loss landscape can be riddled with local minima, wide valleys and other anomalies. Generally, the more challenging the loss space, the more difficult it is to improve upon a given solution.

\subsection{The Search Process}
Given all those spaces, it is perhaps now clearer how the search process is not simply a matter of architecture or size of the neural network. Each of the items mentioned above, e.g. network architecture, different spaces, etc. has a direct impact on the possibility and speed of attaining an acceptable solution. Towards that end, SGD uses information from first-order partial derivatives (gradients) of the loss function to aid it in the search process \cite{Ruder2016AnAlgorithms}, i.e. gradient-following. Despite being computationally costly, gradients can be extremely informative. Over the decades since SGD  was first invented, techniques such as Momentum \cite{Rumelhart1986LearningErrors} were introduced to improve the efficiency of the search process. In addition, by using Backpropagation, and in turn the chain-rule, SGD can update an arbitrary number of weights in a neural network. It may sometimes suffer from complications e.g. vanishing and exploding gradients. There are of course techniques to tackle those, for instance residual connections \cite{He2015DeepRecognition}.

Neuroevolution, on the other hand, does not use derivative information in its search process. Though computationally less-costly, not using derivatives can lead to the optimizer being inefficient and taking more optimization steps than SGD. Moreover, Neurevolution requires the evaluation of a population of solutions before taking a single optimization step. This makes it inherently penalized if one was to compare it directly with single-solution methods on number of evaluations. Without Backpropagation, Neuroevolution suffers when the trained network is relatively large. Each weight in the network becomes an additional dimension in the search space that can take any value within the representation bounds. Without clue or information on how to update each parameter, it essentially becomes a matter of educated guess.

Despite these apparent challenges, a simple genetic algorithm was able to solve Reinforcement Learning tasks using additive Gaussian noise, elitism and no crossover \cite{Such2017DeepLearning}. Note that a population size of 1000 was used. The evolved networks had 4M+ parameters, possibly amongst the largest networks ever evolved. Another result of that work was that Random Search performed surprisingly well, which may perhaps suggest something about the domain itself. In \cite{Salimans2017EvolutionLearning} smaller networks are also evolved for Reinforcement Learning tasks, using Evolution Strategies \cite{Michalewicz1996EvolutionMethods}. Between 720-1440 workers, i.e. populations, were used.

Notably neuroevolution algorithms employed thus far are somewhat linear, as in \cite{Aly2018ExperientialNeuroevolution}. There is no explicit notion of searching multiple regions in the search space. Instead, search would generally be concentrated on one region in hope to transition to more lucrative regions using the perturbation mechanisms. We sought to first address this aspect in the context of Global Optimization problems. The advantage of using Global Optimization functions is that the topology of the solution domain is known. It is possible to visualize optimizer behavior in the solution space, generate insights and determine how best to influence it. Following this line of thought, we describe the MSN metaheuristic in the next section.

\section{Multiple Search Neuroevolution}
A metaheuristic is a strategy that guides the search process \cite{2005ParallelMetaheuristics}. It is not a singular algorithm for a singular problem. We introduce the MSN metaheuristic as a strategy to search multiple regions of the search space effectively. It consists of a set of smaller, locally-aware mechanisms and functions all operating to create a global aggregate behavior. Each of those mechanisms is described in this section, highlighting its function and import. In addition, many new terms and concepts are introduced and utilized. Equivalent terms, where relevant, are mentioned in order to maintain consistency with existing literature and norms. Since the mechanisms are numerous, they are divided into two groups. The primary group contains the core mechanisms without which the MSN scheme can't function. The secondary group contains supplementary mechanisms that are introduced in effort to improve search efficiency under certain conditions.

\subsection{Primary Mechanisms}
\subsubsection{Pool Composition}
The sample pool is the set of candidate samples, i.e. \textit{populations}. The pool size, i.e. number of samples, is always constant. It is extremely important to utilize the given pool as efficiently as possible. This is a major point in our introduction of MSN. In the case of MSN, samples in the pool are the neural networks' parameter vectors. They are initialized according to a weight initialization scheme, e.g. Xavier Normal \cite{Glorot2010UnderstandingNetworks}. After the first optimization step, i.e. \textit{generation}, the pool will consist of the following components: Elite, Anchors, Probes and Blends. Let us introduce each of those in order. First, the Elite is the sample that collected the greatest reward since the optimization process began. Anchors are the highest-rewarded N samples in the current generation. They also need to be at least separated by a certain distance in the search space. This separation mechanism shall be introduced in section \ref{DistanceSection}. From each anchor M probes are spawned. Probes begin as exact clones of their respective anchor after which  each is randomly perturbed.

It is possible that at any arbitrary generation the number of samples sufficiently apart is less than the allotted number of anchors, N. In such cases, the remaining slots are filled with blends, i.e. \textit{crossovers}, which will be described in section \ref{blends section}.

To summarize, the pool is composed of the Elite, N Anchors, M Probes, and Blends, should there are be any open slots. The size of the pool thus must be at least (NxM)+1. The choice of the pool size enables, emphasizes or even disables the Blend mechanism.\\

\subsubsection{Perturbation and Adaptive Integrity}
Perturbation is the primary searching mechanism. It is the equivalent of genetic mutation, principally referring to the injection of noise into the makeup of the network. The magnitude and scope of that noise are determined according to \eqref{search_radius} and \eqref{num_selections}. Equation \eqref{search_radius} describes the Search Radius, i.e. the magnitude of perturbation. We call it Search Radius because it can be thought of as defining the radius of a virtual circle around each anchor, within which probes will be cast in random directions. Generally, as the search radius increases, probes will be casted farther away from an anchor. In the same way that Search Radius defines the magnitude of perturbations, the scope of perturbations is defined by the Number of Selections in \eqref{num_selections}. It determines how many of the weights shall be prone to noise, and how many will be preserved as is.

By examining \eqref{search_radius} and \eqref{num_selections}, it will be noted that they are functions of a variable called \emph{integrity}. It is a single real number in the range [0, 1], and the two equations are designed to have attractive properties in that range. Thus, \emph{integrity} governs the search radius and number of selections. It determines the exploration vs. exploitation aspect on a local scale. Each generation, the algorithm needs to make a decision about \emph{integrity}. Increasing \emph{integrity} makes the search more exploitative and less exploratory, and vice versa. Perhaps this is the single most important decision the algorithm has to take.

The value of \emph{integrity} is reduced, by a fixed amount, when the current generation of samples does not yield a score that is sufficiently better than the previous best score. A parameter in the algorithm defines the minimum accepted percentage of improvement in the reward, we call it \emph{Minimum Entropy}. For example, in the global optimization task, \emph{Minimum Entropy} is set to 1\%. If the current generation does not improve upon the previous generation's reward by at least 1\%, \emph{integrity} is reduced by a fixed \emph{step size}. If it does improve by at least 1\% then the current \emph{integrity} value is maintained.

The Search Radius, a function of \emph{integrity}, is calculated as
\begin{equation}
\label{search_radius}
Search Radius(p) = (tanh((\lambda p)-2.5)+1)*lr, 
\end{equation}
where $\displaystyle p = (1-integrity)$, $\lambda$ and $\displaystyle lr$ are scalar constants. The learning rate, $\displaystyle lr$, scales the function, controlling the bounds of the search radius. The shifted, scaled hyperbolic tangent function has attractive properties in the range [0, 1]. It has an almost-flat slope near 0 and an almost-flat slope near 1. This allows the algorithm to spend more time searching low-energy configurations even as integrity is reduced, where we estimate rewards are more likely. By flattening the slope near 1, it also prevents the algorithm from searching exceedingly high-energy configurations where it is unlikely to find rewards.

The number of selections is calculated as
\begin{equation}
\label{num_selections}
Selections(p) = (\frac{\alpha}{1+\frac{\beta}{p}}), 
\end{equation}
where $\displaystyle p = (1-integrity)$, and $\displaystyle \alpha$ and $\displaystyle \beta$ are scalar constants. In the range [0, 1], this function starts at the origin and progressively becomes flatter as \emph{integrity} is reduced. This saturation limits the number of modifications in the network. Intuitively, making too many adjustments to a neural network in one step is usually unrewarding, especially when the configuration at hand is highly-refined. It is unlikely that changing 70\%, for instance, of a model's weights in one single generation will lead to a higher reward. This is especially the case when searching high-energy configurations, i.e. low \emph{integrity}.The function is designed to saturate, in order to limit a situation where many cycles are wasted searching unprofitable regions.

To summarize, the perturbation mechanism is introduced with its two aspects called search radius and number of selections. The value of \emph{integrity} controls perturbation in order to balance exploration and exploitation locally. Six hyperparameters are defined. Namely they are \emph{step size}, \emph{Minimum Entropy}, $\displaystyle lr$, $\lambda$, $\alpha$ and $\beta$, all of which are determined empirically.\\

\subsubsection{Anchors, Minimum Distance and Probes}
\label{DistanceSection}
Anchors are the N best-performing samples, that are also separated in search space by a certain \emph{Minimum Distance}. For example if there are 5 anchors, those will be the 5 best-performing samples that meet the \emph{Minimum Distance} criterion. Anchors are updated each generation.

The parameter \emph{Minimum Distance} defines the minimum separation requirement for a sample to become a candidate anchor. This guarantees searching different regions in search space. Let us walk through the anchor selection process. After sorting, the best-performing sample is picked as the first anchor. In an ordered reductive process, each sample is then admitted as an anchor if it is separated from the other anchor(s) by at least \emph{Minimum Distance}.

There are many possible distance metrics to choose from such as Euclidean distance. However, we wanted to use a metric that accounts for differences in both magnitude and position (of the weights), between the samples. Thus Canberra distance was chosen, and it is calculated as
\begin{equation}
\label{distance}
d(x,y) = \sum\limits_{i=1}^n (\frac{|x_{i}-y_{i}|}{|x_{i}|+|y_{i}|}) , 
\end{equation}
where $\displaystyle x$, $\displaystyle y$ represent the two parameter vectors, i.e. samples, under examination and $\displaystyle i$ is each element in the vectors.

Finally, from each anchor M exact clones are created, called Probes. Each probe is perturbed, i.e. \textit{mutated}, to produce a different version of the anchor. Thus probes search the local neighborhoods of the anchors. By having multiple anchors, multiple regions are searched in tandem. And by having multiple probes per anchor, multiple local neighborhoods are searched within those regions.

In summary, the processes of choosing anchors and probes are introduced. For anchors, the distance metric is chosen as Canberra distance, given by \eqref{distance} to account for both magnitude and positional differences in the parameter vector. One hyperparameter is introduced called \emph{Minimum Distance}. It defines the least acceptable distance between anchors, and is determined empirically.\\

\begin{figure}[t]
\includegraphics[width=\linewidth]{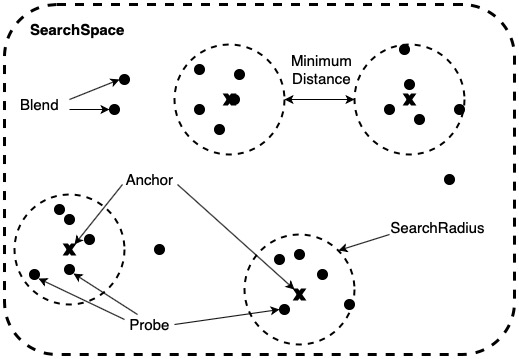}
\caption{Visual outline of primary search mechansims of Multiple Search Neuroevolution in search space.}
\label{system diagram}
\end{figure}

\subsubsection{Blends}
\label{blends section}
Blending, called Crossover in other literature, mechanism combines randomly-chosen weights from two components to yield a blend. The first component is always one of the anchors, picked at random. The second component can be any sample in the pool. The first component is cloned to form the basis. A number of weights from the second component replace their counterparts in the basis. This number is calculated by \eqref{num_selections}, introduced in the previous section.

Blending potentially allows the exploration of regions outside the main mechanism of perturbation and its constraints. By being based on anchors, blends attempt to extend the actively searched area in the search space. This exploration behavior can be emphasized or discouraged by increasing or reducing the number of slots allotted to blends in the pool.

In addition, it is possible that not enough samples are sufficiently apart in the search space to fill the N spots allotted to anchors. This would also mean that not enough probes would be made, and thus many slots in the pool would remain unfilled. Blends fill all those slots, helping to reintroduce diversity into the pool. With diversity, distance between samples increases. In consequence, the allotted slots for anchors can be filled again.

In summary, this section introduced blends and how they are picked. Blends attempt to explore regions beyond the constraints of perturbation. Also, they are important to fill open slots in the pool and maintain diversity. A visual overview of the system is given in Fig. \ref{system diagram}.\\

\subsubsection{Elitism}
Elitism is well-known throughout, we mention it for completeness. There are two sorts of elites, a generational elite, and a historical elite. The generational elite is the best-performing, i.e. highest-rewarded, sample of the current generation. The historical elite is the best-performing sample across all generations. In MSN, the generational elite is always picked as the first of the Anchors. The historical elite is simply called the Elite. It is preserved as-is, without being subject to perturbation or blends, across generations unless another better-performing sample replaces it. The Elite is called upon whenever the mechanism of Backtracking is invoked, which will be presented in section \ref{backtracking}.

\subsection{Secondary Mechanisms}
\subsubsection{Backtracking}
\label{backtracking}
Without sufficient improvement in the reward signal, \emph{integrity} is reduced. This makes perturbations and blends more potent, searching higher-energy configurations and encouraging exploration. This is a uni-directional behavior, and MSN will keep reducing integrity (to its limit of 0) until an improvement is achieved. This has a number of disadvantages, however. First, it may just be the case that the lucrative areas are actually in low-energy configurations. Due to the probabilistic nature of the algorithm, it is quite possible the algorithm "missed". Second, with low integrity the samples become "hot" from applying high-magnitude perturbations on a large scope, and the weights start exploding.

Backtracking is a mechanism that resets \emph{integrity}. It is triggered when MSN reduces \emph{integrity} for X consecutive generations, where X is a parameter called \emph{Patience}. When it is triggered, backtracking resets the \emph{integrity} value back to maximum, and inserts the Elite as an Anchor in the pool. This accomplishes two things. First, by resetting \emph{integrity}, the algorithm will search low-energy regions once more. Second, inserting the elite as an anchor, and spawning its probes with maximum \emph{integrity}, helps "cool down" the entire pool. This is because the weights of the elite were not subject to perturbations. Thus, the search returns to low-energy configurations.

In summary, if \emph{integrity} is consistently decreased without improvement, the search process may need to be reset. Backtracking is mechanism to accomplish this. One hyperparameter is introduced, called \emph{Patience}. It determines how many generations the algorithm should wait before backtracking, and it is determined empirically.\\

\subsubsection{Radial Expansion}
Radial expansion increases the learning rate $\displaystyle lr$ and $\displaystyle \alpha$ from \eqref{search_radius} and \eqref{num_selections} by a fixed percentage called \emph{Expansion Factor}. Let us explain the reasoning behind this. Even at lowest \emph{integrity}, there is a limit to how far a probe can be cast from its anchor, as defined by those equations. If that limit is too constrictive for the task, search efficiency will suffer for two reasons. The first: it can lead to the full number of anchors not being utilized, as the samples are not far enough from each other. The second: it can increase the number of exploratory search iterations since the algorithm is too conservative. Radial expansion is triggered whenever the number of anchors becomes less than the allotted number, N.

In summary, the mechanism of Radial Expansion decreases the constrictions of local search. It is an attempt to utilize the complete number of anchors. One hyperparameter is introduced, called \emph{Expansion Factor}, and is determined empirically.

\section{Implementation and Results}
\begin{table*}[t]
\caption{Experimental results of optimization algorithms on typical global optimization functions. A few entries are missing because the corresponding library implementations are unavailable. The Speedup column shows the improvement in optimization steps taken when using MSN metaheuristic against other algorithms, discounting those that did not converge.}
\label{results on common}
\begin{centering}

\begin{tabular}{c|c|c|c|c|c|c|c|c|c}
\hline
\multirow{2}{*}{\bf Function} & \multicolumn{8}{c}{\bf Number of Optimization Steps}\\
& MSN & ES & PSO & DE & Simulated & FEM & PGPE & Random & Speedup\\
& & & & & Annealing & & & Search &
\\ \hline \\
Ackley		&17 &36 &117 &659 &5000+ &287 &152 &4149 &2.1 - 244X\\
Rastrigin	&49 &2020 &418 &632 &2368 &2389 &421 &3074 &8.5 - 63X\\
Rosenbrock	&20 &67 &730 &2415 &2398 &- &- &- &3.3 - 120X\\
Schwefel	&113 &2019 &492 &2310 &5000+ &- &- &- &4.3 - 20X\\
\end{tabular}

\end{centering}
\end{table*}
For all experiments in this paper, a pool size of 50 samples is used. This is a remarkably low number, compared to other works using evolutionary algorithms on neural networks such as \cite{Such2017DeepLearning} and \cite{Salimans2017EvolutionLearning} using more samples by one or two orders of magnitude. This choice, we believe, emphasizes efficiency. Thus, with a relatively small pool, the goal is to converge in the upcoming tests in as little iterations as possible. The algorithm is implemented using the PyTorch framework \cite{Paszke2017AutomaticPyTorch}. The implementation is our intellectual property, and there is no plan to publicly release the software foreseeable future. Despite this, an open-source version is being actively developed and should be released in due course. Moreover, the information contained in this paper should be sufficient for any developer to implement the algorithm. 

The training process with MSN can be described as follows. Since MSN is population-based, a pool of networks is being evolved simultaneously. At the beginning, each network is initialized according to a weight initialization scheme, we use Xavier Normal. In each iteration, i.e. optimization step, a query of the environment is conducted by each network. The networks each take the same input and produce their own output. Then the loss/reward/cost signal is computed and fed to the algorithm. It informs the algorithm about the quality of each network's output. Finally, MSN adjusts the weights of the networks and the process is repeated.

The experiments run on an Nvidia DG-X desktop computer, featuring four Nvidia Titan V GPUs. In the Global Optimization task, the experiments run on CPU and a single GPU. In the MNIST hand-written digit classification task, the experiments run on CPU and the four GPUs for data parallelization. The models are copied into each GPU and inference is conducted by dividing the training set/batch into four chunks, computing loss and aggregating the result. This is only to speed wall-clock time, and does not affect the performance of the algorithm or the results of inference. For all experiments a pool size of 50, with 4 anchors, and 8 probes per anchor are used.

\subsection{Task 1: Global Optimization}
We test MSN on two sets of Global Optimization functions. The first is a standard group and part of the BBOB challenge suite \cite{Tusar2016COCO:Suite}, a measure introduced in 2016. Those functions are also commonly implemented in different Python libraries, and thus allow us to compare MSN to other evolutionary algorithms. The second set is a special group of functions that are less common, and neither a part of the BBOB nor implemented in evolutionary libraries. Thus, A comparison with other algorithms on that set was not possible. Nonetheless, the challenges they pose are unique due to their irregular topographies. We thought it could be helpful to measure the performance of the algorithm against them.

The optimization task is straightforward and simple. Starting from a random location (x, y) in the search space, the algorithm needs to find the global optimum of the 2-dimensional function, or approximate it. If the algorithm is within 0.06 of the global optimum value, the search terminates as a success. For example, if the global optimum is 0, then as soon as the algorithm reaches 0.06 or less it terminates. The number of optimization steps taken until termination is recorded, and is the measure of performance in the experiments. The lower the number of optimization steps needed to converge, the better the algorithm performs. The algorithms also terminate automatically if the number of steps taken exceed 5000. It is unlikely that the algorithm would converge beyond that point, and for reasons of comparison, it would not mean much if it did either.

In Task 1, MSN optimizes a neural network model with a single hidden-layer of size 128. The network takes two real numbers (x, y) as the origin, picked uniformly from the observation space and remain constant throughout that experiment. The uniform distribution's limits differ for each function. The network outputs two real numbers, its prediction of the coordinates of the global optimum. Other evolutionary algorithms don't optimize neural networks, they operate directly on the problem. The implementations of those algorithms are found in the Inspyre and PyBrain evolutionary libraries \cite{Garrett2017Inspyred:Python} \cite{Schaul2010PyBrain}. The default parameters are used. Generally, parameters in standard library implementations are either directly copied from the algorithm's paper or picked to suite a wide set of problems. They are certainly not adversarial. For a fair comparison, however, we did not tune our algorithm, MSN, parameters either. We determine suitable values, empirically, and then use that exact same set across all experiments.

For every function, the optimization experiment is repeated five times per algorithm. The average number of optimization steps, i.e. generations, until termination is recorded. Limits for the functions are given as a single pair for symmetrical limits, and in the form (x-start, x-limit, y-start, y-limit) for asymmetrical limits. They are: [-5,5] for Ackley, [-5.2,5.2] for Rastrigin, [-2,2] for Rosenbrock, [-500,500] for Schwefel, [-15,-5,-3,3] for Bukin N. 6, [-20,20] for Easom and [-512,512] for Eggholder. The same limits are used in all experiments. The definitions of the functions can be found in \cite{SonjaSurjanovicDerekBingham2013OptimizationDatasets}.

The first group of consisted of the Ackley, Rastrigin, Rosenbrock and Schwefel optimization functions. The tested algorithms are Evolution Strategies \cite{Michalewicz1996EvolutionMethods} (ES), Particle Swarm Optimization (PSO) \cite{Kennedy2011ParticleOptimization}, Differential Evolution (DE) \cite{Qin2009DifferentialOptimization}, Simulated Annealing \cite{VanLaarhoven1987SimulatedAnnealing}, Fitness-Maximization Expectation (FEM) \cite{Wierstra2008FitnessMaximization}, Policy Gradients with Parameter Exploration (PGPE) \cite{Sehnke2010Parameter-exploringGradients} and Random Search. For some of those algorithms, not all the functions were available in their libraries for testing. However, at least two functions were tested in each case.

The mode of comparison for the algorithms is the number of optimization steps performed. All of them had a population/swarm of size 50. It remained to be seen how well each will use the available resources to guide the optimization process. Table \ref{results on common} presents the results of the experiments from the first group. The results warrant some analysis. Some methods fared well on some functions but struggled on others, such as Evolution Strategies. some methods struggled consistently, such as Simulated Annealing which failed to converge on two occasions. Unlike the findings of \cite{Such2017DeepLearning}, Random Search performed poorly. As expected, it may be a subject of the extremely limited pool size we use, as well as the problem domain. The strongest competitors were Evolution Strategies, PSO and PGPE. The consistent best-performer, however, was MSN.

The improvement to using MSN compared to the other algorithms is calculated and presented in the Speedup column. At least, MSN reduced the optimization steps by 2X, and at best by 244X. As expected, the Schwefel function took the longest for MSN to solve. 

\begin{figure}[t]
\begin{subfigure}{0.5\textwidth}
\includegraphics[width=0.9\linewidth, height=5cm]{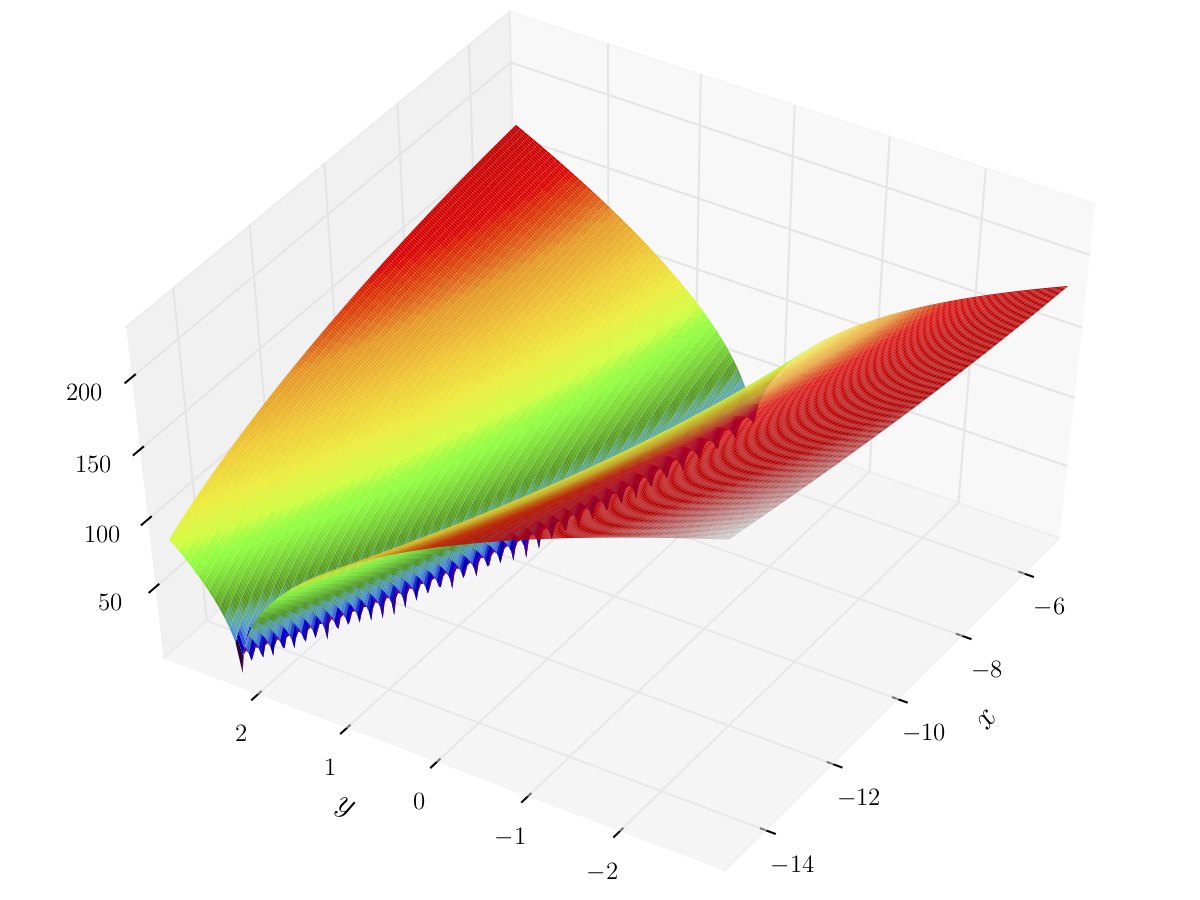} 
\caption{Bukin N.6 function}
\label{fig:bukin}
\end{subfigure}
\begin{subfigure}{0.5\textwidth}
\includegraphics[width=0.9\linewidth, height=5cm]{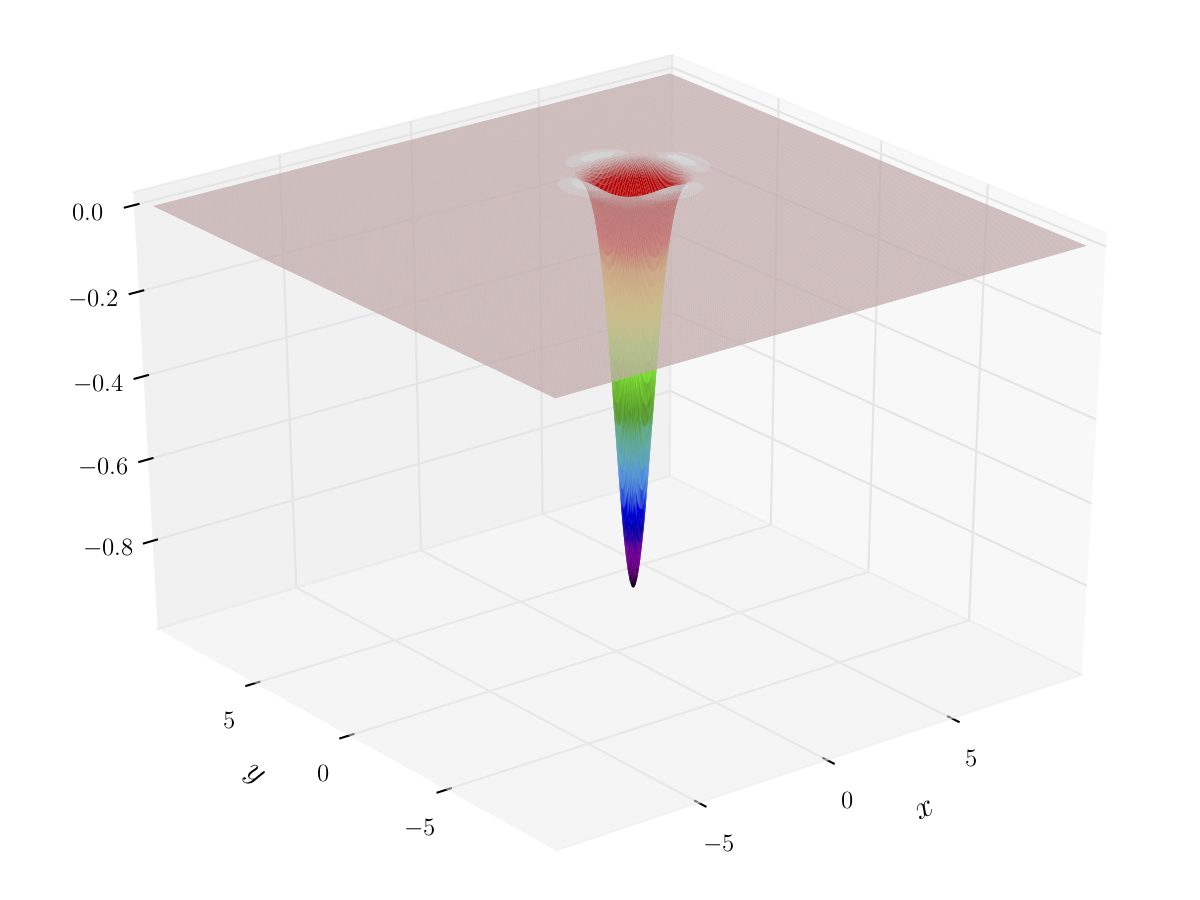} 
\caption{Easom function}
\label{fig:easom}
\end{subfigure}
\begin{subfigure}{0.5\textwidth}
\includegraphics[width=0.9\linewidth, height=5cm]{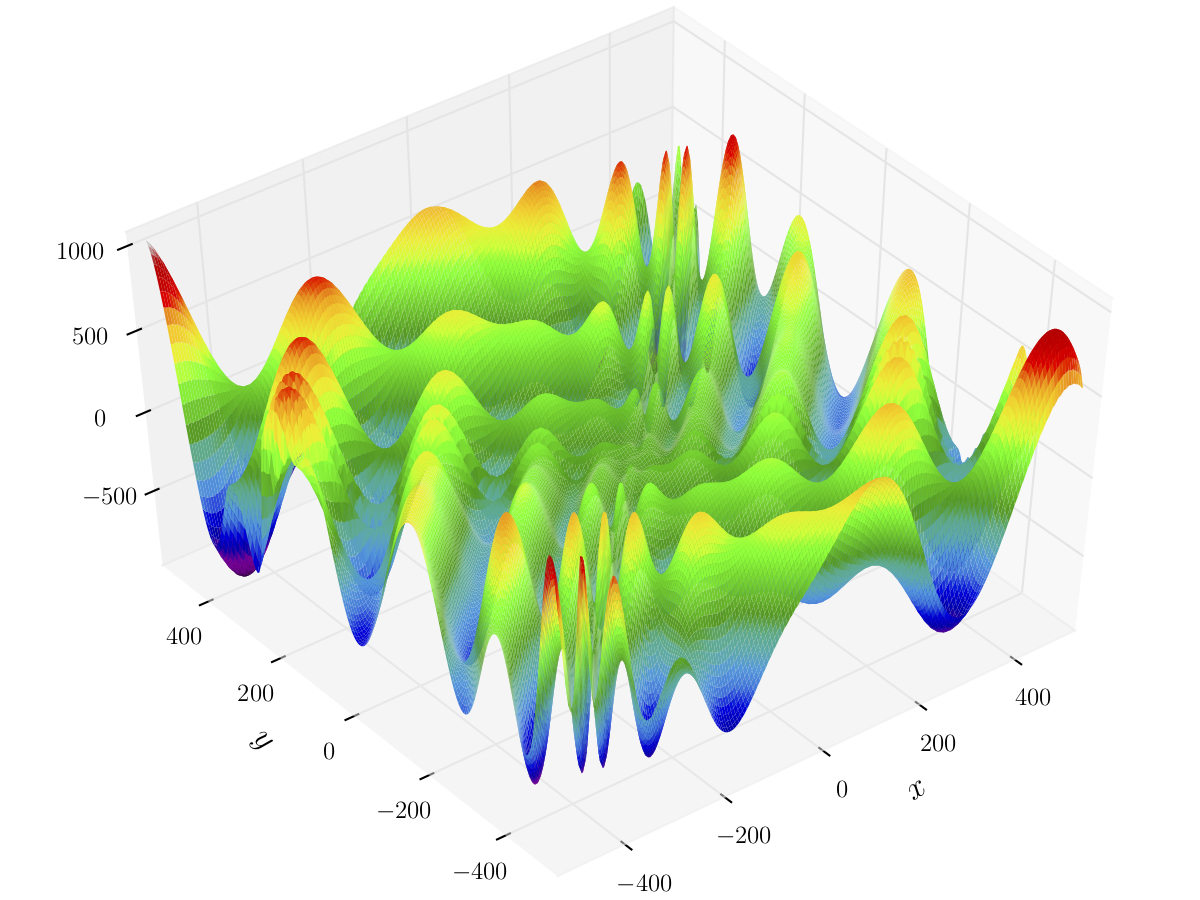} 
\caption{Eggholder function}
\label{fig:eggholder}
\end{subfigure}
\caption{3D plots of the special group of global optimization functions. Credit: \cite{GaortizgFile:BukinWikipedia}, \cite{GaortizgFile:EasomWikipedia}, \cite{GaortizgFile:EggholderWikipedia}}
\label{fig:function plots}
\end{figure}

The special group of functions is composed of the Bukin N. 6, Easom and Eggholder functions. Let us examine each in turn. The Bukin N. 6 function has a unique feature of an extremely narrow valley where the global optimum lies. This poses a challenge for the exploitation aspect of an optimization algorithm. If an algorithm takes too large optimization steps, it is unlikely to find that lucrative strip. The Easom function is practically flat everywhere except a thin spike, where the global optimum lies. The algorithm needs to explore as fast as possible the search space, and then once it finds the spike, to being exploitation and travel towards the global optimum. It showcases a balance between strong exploration and exploitation. Finally, the Eggholder function is extremely irregular and adversarial in its shape. Its global optimum is actually in a corner. The optimizing algorithm needs to overcome the large number of deep local minima, and keep searching for the global optimum. Of all the functions, it has the largest solution space. A visual representation of the functions' landscape is given in Fig. \ref{fig:function plots}.

The results of the experiments on the group of special optimization functions is given in Table \ref{results on special}. In all cases, MSN converged in a relatively low number of iterations. Particularly in the case of the Easom function. By searching multiple regions separated by a distance metric, the algorithm is naturally exploration-oriented. The task of traversing a smooth-but-flat surface did not strain the algorithm. Similarly, by using the Probe mechanism, it is also naturally exploitative. Thus finding and exploiting the extremely narrow valley in the Bukin N. 6 function was not taxing. The most challenging function was, as expected, Eggholder. Its highly irregular landscape and having its global optimum in a corner proved challenging. Recall also that its solution space is the largest.

\begin{table}[t]
\caption{Experimental results of running MSN metaheuristic on optimization functions with special properties.}
\label{results on special}
\begin{center}
\begin{tabular}{c|c|c|c}
\hline
{\bf Function} & Bukin N. 6 &  Easom & Eggholder
\\ \hline
Number of Optimization Steps & 28 & 9 & 170
\\ \hline
\end{tabular}
\end{center}
\end{table}

\subsection{Task 2: Image Classification}
For Task 2, a convolutional neural network (CNN) is trained to solve the MNIST hand-written digit classification problem. It is a standard entry-level problem where the goal is to correctly classify 28x28 greyscale images according to the number they feature. The CNN model consists of four convolutional layers of size 32, 64, 128 and 128 respectively with stride 2 and max-pooling in between, followed by a single fully-connected layer of size 512. Parametric ReLu non-linearity in PyTorch was used. The model has 4.7M parameters. 

The following set of conditions were imposed during the experiments. Since MSN is population-based, a pool of networks are being evolved simultaneously. This stretches inference time linearly. To speed up computations, leveraging NVIDIA GPU Tensor Cores, half-precision floats (FP16) are used. Furthermore, only a subset of 2000 randomly-picked images (3.33\%) of the MNIST training set is used. However, the entire validation set is used. We set the termination condition to be a loss of 0.15. Given only a subset of the training set, demanding further improvement is likely to introduce a cycle of diminishing returns. All this comes at a cost of not reaching the best possible performance on the task, but that is not the concern of this work. The goal from this task is to assess the suitability of MSN to optimize relatively large neural networks with 10\textsuperscript{6} parameters.

As a baseline, the same CNN model is trained under the same conditions with SGD. Note that SGD uses mini-batch training. It updates the weights after inference on every mini-batch. By comparison, MSN does not use mini-batch training. It performs a weight update after inference on the entire training set (2000 images). For that reason, comparing SGD to MSN on the number of absolute inferences would be inherently flawed. Moreover, recall that these two belong to different families of optimization algorithms, one is a single-solution while the other is population-based. The experiment is repeated five times and the mean is reported in Table \ref{results on mnist}.

From Table III, MSN takes 2333 generations to converge to the target training loss. This corresponds to ~90\% validation accuracy. In general, MSN is able to train the relatively large network, using a pool of only 50 samples. Compared to Task 1, the observation space is larger by two orders of magnitude. The search space also is larger by four orders of magnitude. Despite this, MSN takes only an order of magnitude more steps to converge for MNIST than the Eggholder function.

\begin{table}[t]
\caption{Experimental results of running optimization algorithms on MNIST dataset, using 3.33\% of its training set.}
\label{results on mnist}
\begin{center}
\begin{tabular}{c|c|c|c}
\hline
{\bf Algorithm} &MSN &SGD &Speedup
\\ \hline
Number of Optimization Steps &2333 &320 &-7.3X
\\ \hline
Validation accuracy (\%) &90 &90
\\ \hline
\end{tabular}
\end{center}
\end{table}

\begin{figure}[t]
\begin{center}
\begin{tikzpicture}
\begin{axis}[xlabel=Optimization steps, ylabel=Training Loss]
\addplot [thick, smooth, each nth point = 100] table[x=Generations,y=Loss,col sep=comma, mark=none]{data1.csv};
\addplot [thick, smooth, blue] table[x=inference,y=loss,col sep=comma, mark=none]{sgd_mnist_loss.csv};
\end{axis}
\end{tikzpicture}
\caption{Training progress over optimization steps from a sample of the experiments performed on Task 2. Experiments terminate once training loss of 0.1 is achieved. As the training progresses, the gap between MSN and SGD notably widens. The Black trace is MSN and Blue is SGD.}
\label{mnist sample figure}
\end{center}
\end{figure}
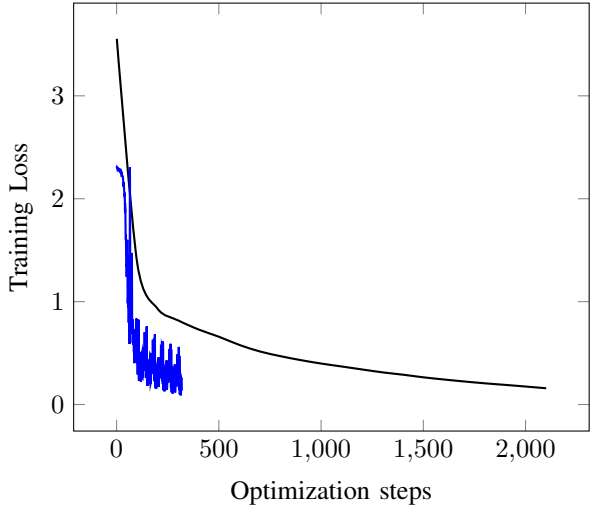

On the other hand, SGD is able to reach the target training loss taking only 320 steps. The speedup factor is -7X, since instead of being faster, our algorithm is slower than the baseline. That MSN would be slower than SGD is not surprising. Recall that SGD utilizes first-order partial derivatives (gradients) to guide its search process. On the other hand MSN does not have access to such information, being derivative-free. Remarkably, however, it is still within an order of magnitude of baseline.

Fig. \ref{mnist sample figure} shows sample of training progress over optimization steps. It is drawn using experimental data from one of the five trials performed. In that particular experiment, the target training loss was achieved at step 320 for SGD, shown in Blue, and 2161 for MSN, shown in Black. Our algorithm was 6.75X slower than baseline in that case. The figure also showcases how swiftly training loss decreases in the early stages, for both algorithms. However, after going below a training loss of 1, SGD continued to improve at a slightly slower rate while MSN started to plateau. This perhaps indicates a limitation in the exploitative nature of MSN for large search spaces. 

The figure also suggests that SGD has not fully converged. If allowed more steps, it would have likely reduced the loss further. For MSN, the algorithm seemingly started to plateau at step 1,500 and every further step achieved diminished returns. As such, the gap between MSN and SGD widens as training loss is reduced.

Using only 50 populations, the CNN trained in Task 2 is by far the largest trained using derivative-free methods. While not state-of-the-art, the performance of the network meets the predefined target. To the best of our knowledge no other work managed to train relatively large neural networks under such constrained conditions. In comparison to recent publications \cite{Such2017DeepLearning} and \cite{Salimans2017EvolutionLearning} that attempt to use derivative-free methods to train neural networks, we use fewer populations by 14-28X. The task domains were different, however.

\section{Conclusion}
Drawing upon the limitations of derivative-based single-solution optimization methods, the paper introduced a new metaheurisitc called Multiple Search Neuroevolution (MSN). It is a derivative-free population-based strategy that guides the optimization process of deep neural networks. Its ensemble of mechanisms is presented in detail, divided into two groups.

An implementation of the MSN metaheurisitc is tested on two tasks. In both, MSN optimizes a pool of neural networks to solve the task. The first task was to find the global optimum for groups of common and special global optimization functions. In solving the problems, MSN reduced the number of optimization steps by 2-244 X, compared to baseline evolutionary algorithms. Featuring nine empirically-derived hyperparameters, however, MSN is certainly not as simple as the baseline. Thus there is a clear trade-off between speedup and complexity.

The second task was to reach the target training loss on the MNIST hand-written digit classification dataset. Using 3.33\% of the training set, MSN required 7X more optimization steps then the baseline algorithm (SGD) to reach the termination condition. This was anticipated since SGD utilizes gradient information to guide the search process, while MSN is derivative-free. The results further suggest a limitation in the exploitative mechanisms. No doubt under such a constrained pool size, it is challenging to balance exploration and exploitation. That being said, perhaps more attention is due to the exploitative aspect.

The study has some limitations. First, only two-dimensional functions were used. In future work, testing higher-dimensional functions would be helpful. Second, only algorithms available in standard evolutionary libraries were used as baseline. In future implementations, it would be helpful to compare MSN to more competitive evolutionary algorithms. Finally, since MSN is presented as a collective, and the role of each mechanism separately is not quantifiable. Performing an ablation study would be helpful to investigate the individual effect of each search mechanism, for different types of problems.

Nonetheless, the results show that there is credence to the mechanisms introduced. Consistently outperforming the other evolutionary algorithms in Task 1 is a significant indicator. The model in Task 2 featured 4.7M parameters, which is a significant search space for evolutionary methods. In addition, only 50 populations were used. Yet, MSN was able to train to 90\% validation accuracy using 3.33\% of the available training set.

\section*{Acknowledgment}
We thank Vik Goel and Qiyang Li for their insightful comments and discussions.

\bibliography{ea_congress_refs}

\begin{thebibliography}{10}

\bibitem{Liu2016Ssd:Detector}
W.~Liu, D.~Anguelov, D.~Erhan, C.~Szegedy, S.~Reed, C.-Y. Fu, and A.~C. Berg,
  ``{Ssd: Single shot multibox detector},'' in {\em European conference on
  computer vision}, pp.~21--37, 2016.

\bibitem{Lenz2015DeepGrasps}
I.~Lenz, H.~Lee, and A.~Saxena, ``{Deep learning for detecting robotic
  grasps},'' {\em The International Journal of Robotics Research}, vol.~34,
  no.~4-5, pp.~705--724, 2015.

\bibitem{Bahdanau2014NeuralTranslate}
D.~Bahdanau, K.~Cho, and Y.~Bengio, ``{Neural machine translation by jointly
  learning to align and translate},'' {\em arXiv preprint arXiv:1409.0473},
  2014.

\bibitem{CauchyMethodeSimultanees}
A.~Cauchy, ``{M{\'{e}}thode g{\'{e}}n{\'{e}}rale pour la r{\'{e}}solution des
  systemes d'{\'{e}}quations simultan{\'{e}}es},'' {\em cs.xu.edu}.

\bibitem{Robbins1951AMethod}
H.~Robbins and S.~Monro, ``{A Stochastic Approximation Method},'' {\em The
  Annals of Mathematical Statistics}, vol.~22, pp.~400--407, 9 1951.

\bibitem{Kingma2014Adam:Optimization}
D.~P. Kingma and J.~Ba, ``{Adam: A Method for Stochastic Optimization},'' 12
  2014.

\bibitem{WERBOS1974BeyondScience}
{WERBOS} and {P.}, ``{Beyond Regression : New Tools for Prediction and Analysis
  in the Behavior Science},'' {\em Unpublished Doctoral Dissertation, Harvard
  University}, 1974.

\bibitem{Ioffe2015BatchShift}
S.~Ioffe and C.~Szegedy, ``{Batch Normalization: Accelerating Deep Network
  Training by Reducing Internal Covariate Shift},'' 2 2015.

\bibitem{Luong2015EffectiveTranslation}
M.-T. Luong, H.~Pham, and C.~D. Manning, ``{Effective approaches to
  attention-based neural machine translation},'' {\em arXiv preprint
  arXiv:1508.04025}, 2015.

\bibitem{Xu2015ShowAttention}
K.~Xu, J.~Ba, R.~Kiros, K.~Cho, A.~Courville, R.~Salakhudinov, R.~Zemel, and
  Y.~Bengio, ``{Show, attend and tell: Neural image caption generation with
  visual attention},'' in {\em International conference on machine learning},
  pp.~2048--2057, 2015.

\bibitem{Sorensen2015Metaheuristics-theExposed}
K.~S{\"{o}}rensen, ``{Metaheuristics-the metaphor exposed},'' {\em
  International Transactions in Operational Research}, vol.~22, pp.~3--18, 1
  2015.

\bibitem{2005ParallelMetaheuristics}
E.~Alba, ed., {\em {Parallel Metaheuristics}}.
\newblock Hoboken, NJ, USA: John Wiley {\&} Sons, Inc., 8 2005.

\bibitem{Ronald1994GeneticControl}
E.~Ronald and M.~Schoenauer, ``{Genetic lander: An experiment in accurate
  neuro-genetic control},'' pp.~452--461, Springer, Berlin, Heidelberg, 1994.

\bibitem{Miikkulainen2002EvolvingTopologies}
K.~O.~S. Miikkulainen and Risto, ``{Evolving Neural Networks Through Augmenting
  Topologies},'' {\em Evolutionary Computation}, vol.~10, no.~2, pp.~99--127,
  2002.

\bibitem{Such2017DeepLearning}
F.~P. Such, V.~Madhavan, E.~Conti, J.~Lehman, K.~O. Stanley, and J.~Clune,
  ``{Deep neuroevolution: genetic algorithms are a competitive alternative for
  training deep neural networks for reinforcement learning},'' {\em arXiv
  preprint arXiv:1712.06567}, 2017.

\bibitem{Goodfellow2014GenerativeNets}
I.~Goodfellow, J.~Pouget-Abadie, M.~Mirza, B.~Xu, D.~Warde-Farley, S.~Ozair,
  A.~Courville, and Y.~Bengio, ``{Generative adversarial nets},'' in {\em
  Advances in neural information processing systems}, pp.~2672--2680, 2014.

\bibitem{Finn2016DeepLearning}
C.~Finn, X.~Y. Tan, Y.~Duan, T.~Darrell, S.~Levine, and P.~Abbeel, ``{Deep
  spatial autoencoders for visuomotor learning},'' in {\em 2016 IEEE
  International Conference on Robotics and Automation (ICRA)}, pp.~512--519,
  2016.

\bibitem{Mnih2013PlayingLearning}
V.~Mnih, K.~Kavukcuoglu, D.~Silver, A.~Graves, I.~Antonoglou, D.~Wierstra, and
  M.~Riedmiller, ``{Playing atari with deep reinforcement learning},'' {\em
  arXiv preprint arXiv:1312.5602}, 2013.

\bibitem{Goodfellow2016DeepLearning}
I.~Goodfellow, Y.~Bengio, A.~Courville, and Y.~Bengio, {\em {Deep learning}},
  vol.~1.
\newblock MIT Press, 2016.

\bibitem{Salimans2017EvolutionLearning}
T.~Salimans, J.~Ho, X.~Chen, S.~Sidor, and I.~Sutskever, ``{Evolution
  strategies as a scalable alternative to reinforcement learning},'' {\em arXiv
  preprint arXiv:1703.03864}, 2017.

\bibitem{Ruder2016AnAlgorithms}
S.~Ruder, ``{An overview of gradient descent optimization algorithms},'' 9
  2016.

\bibitem{Rumelhart1986LearningErrors}
D.~E. Rumelhart, G.~E. Hinton, and R.~J. Williams, ``{Learning representations
  by back-propagating errors},'' {\em Nature}, vol.~323, pp.~533--536, 10 1986.

\bibitem{He2015DeepRecognition}
K.~He, X.~Zhang, S.~Ren, and J.~Sun, ``{Deep Residual Learning for Image
  Recognition},'' 12 2015.

\bibitem{Michalewicz1996EvolutionMethods}
Z.~Michalewicz, ``{Evolution strategies and other methods},'' in {\em Genetic
  Algorithms+ Data Structures= Evolution Programs}, pp.~159--177, Springer,
  1996.

\bibitem{Aly2018ExperientialNeuroevolution}
A.~Aly and J.~B. Dugan, ``{Experiential Robot Learning with Accelerated
  Neuroevolution},'' 8 2018.

\bibitem{Glorot2010UnderstandingNetworks}
X.~Glorot and Y.~Bengio, ``{Understanding the difficulty of training deep
  feedforward neural networks},'' in {\em Proceedings of the thirteenth
  international conference on artificial intelligence and statistics},
  pp.~249--256, 2010.

\bibitem{Paszke2017AutomaticPyTorch}
A.~Paszke, S.~Gross, S.~Chintala, G.~Chanan, E.~Yang, Z.~DeVito, Z.~Lin,
  A.~Desmaison, L.~Antiga, and A.~Lerer, ``{Automatic differentiation in
  PyTorch},'' in {\em NIPS-W}, 2017.

\bibitem{Tusar2016COCO:Suite}
T.~Tusar, D.~Brockhoff, N.~Hansen, and A.~Auger, ``{COCO: The Bi-objective
  Black Box Optimization Benchmarking (bbob-biobj) Test Suite},'' 4 2016.

\bibitem{Garrett2017Inspyred:Python}
A.~Garrett, ``{inspyred: Bio-inspired Algorithms in Python},'' 6 2017.

\bibitem{Schaul2010PyBrain}
T.~Schaul, J.~Bayer, D.~Wierstra, Y.~Sun, M.~Felder, F.~Sehnke,
  T.~R{\"{u}}ckstie{\ss}, and J.~Schmidhuber, ``{PyBrain},'' {\em Journal of
  Machine Learning Research}, vol.~11, pp.~743--746, 2010.

\bibitem{SonjaSurjanovicDerekBingham2013OptimizationDatasets}
{Sonja Surjanovic; Derek Bingham}, ``{Optimization Test Functions and
  Datasets},'' 2013.

\bibitem{Kennedy2011ParticleOptimization}
J.~Kennedy, ``{Particle swarm optimization},'' in {\em Encyclopedia of machine
  learning}, pp.~760--766, Springer, 2011.

\bibitem{Qin2009DifferentialOptimization}
A.~K. Qin, V.~L. Huang, and P.~N. Suganthan, ``{Differential evolution
  algorithm with strategy adaptation for global numerical optimization},'' {\em
  IEEE transactions on Evolutionary Computation}, vol.~13, no.~2, pp.~398--417,
  2009.

\bibitem{VanLaarhoven1987SimulatedAnnealing}
P.~J.~M. Van~Laarhoven and E.~H.~L. Aarts, ``{Simulated annealing},'' in {\em
  Simulated annealing: Theory and applications}, pp.~7--15, Springer, 1987.

\bibitem{Wierstra2008FitnessMaximization}
D.~Wierstra, T.~Schaul, J.~Peters, and J.~Schmidhuber, ``{Fitness expectation
  maximization},'' in {\em International Conference on Parallel Problem Solving
  from Nature}, pp.~337--346, 2008.

\bibitem{Sehnke2010Parameter-exploringGradients}
F.~Sehnke, C.~Osendorfer, T.~R{\"{u}}ckstie{\ss}, A.~Graves, J.~Peters, and
  J.~Schmidhuber, ``{Parameter-exploring policy gradients},'' {\em Neural
  Networks}, vol.~23, no.~4, pp.~551--559, 2010.

\bibitem{GaortizgFile:BukinWikipedia}
{Gaortizg}, ``{File:Bukin function 6.pdf - Wikipedia}.''

\bibitem{GaortizgFile:EasomWikipedia}
{Gaortizg}, ``{File:Easom function.pdf - Wikipedia}.''

\bibitem{GaortizgFile:EggholderWikipedia}
{Gaortizg}, ``{File:Eggholder function.pdf - Wikipedia}.''

\end{thebibliography}
\bibliographystyle{ieeetr}

\end{document}